# The MeSH-gram Neural Network Model: Extending Word Embedding Vectors with MeSH Concepts for UMLS Semantic Similarity and Relatedness in the Biomedical Domain

Saïd Abdeddaïm[a], Sylvestre Vimard[a], Lina F. Soualmia[a]

*[a] Normandie Univ., UNIROUEN, UNIHAVRE, INSA Rouen, LITIS, F-76000, Rouen, France,*

**Abstract**

*Eliciting semantic similarity between concepts remains a challenging task. Recent approaches founded on embedding vectors have gained in popularity as they risen to efficiently capture semantic relationships. The underlying idea is that two words that have close meaning gather similar contexts. In this study, we propose a new neural network model named "MeSH-gram" which relies on a straightforward approach that extends the skip-gram neural network model by considering MeSH (Medical Subject Headings) descriptors instead words. Trained on publicly available PubMed/MEDLINE corpus, MesSH-gram is evaluated on reference standards manually annotated for semantic similarity. MeSH-gram is first compared to skip-gram with vectors of size 300 and at several windows' contexts. A deeper comparison is performed with twenty existing models. All the obtained results of Spearman's rank correlations between human scores and computed similarities show that MeSH-gram (i) outperforms the skip-gram model, and (ii) is comparable to the best methods but that need more computation and external resources.*

## Introduction

Eliciting semantic similarity and relatedness between concepts is a major issue in the biomedical domain. Different measures have been proposed the last decades [1]. Those measures quantify the degree to which two concepts are similar. They either rely on knowledge-based approaches using ontologies and terminologies, or corpus-based approaches which are founded on distributional statistics (e.g. literature-based drug discovery [2-5]). Several clinical applications of importance rely on semantic similarity and relatedness [6], such as biomedical information extraction and retrieval, clinical decision support, or disease prediction. For instance, biomedical information extraction and retrieval is improved by including semantically related terms and concepts [7-10]. The same approaches are used in the task of summarizing Electronic Health Records [11,12] and in document clustering [13]. The prediction of disease-causing genes and disease prediction from similar genes [14,15] rely on the identification of similar dideases [16], or genes [17]. Other applications include drug re-purposing [18,19] and drug interaction [20].

The recent approaches that have given better results in semantic similarity and relatedness measures are founded on word embedding vectors computed by neural networks. Indeed, such architectures implemented initially by word2vec [21], have gained in popularity in the biomedical domain as they risen to efficiently capture semantic similarity and relatedness relationships between words and concepts [22-27]. Word embeddings is based on neural network language modeling where words are mapped to fixed-dimension vectors of real numbers. The similarity between words can thus be measured by the (cosine) similarity between vectors that are constructed over a training corpus. All co-occurrences of a word and its neighbors (i.e. contexts) within a predefined window size are considered. The idea behind those representation learning approaches is that two words that have close meaning have generally similar contexts [28]. For example, the words "Epilepsy" and "Convulsion" will both have "Brain" and "Mind" as neighbors.

word2vec developed by Mikolov et al.[21] is a neural network language model that learns word vectors that either maximizes the probability of a word given the surrounding context, referred to as the CBOW approach (Continuous Bag Of Words) approach, or to maximize the probability of the context given a word, referred to as the skip-gram approach.

In this study we propose a new method, named "MeSH-gram", which relies on a straightforward approach: it computes the word vectors by only using the MeSH (Medical Subject Headings) descriptors that are already included in the MEDLINE/PubMed corpus. The MeSH-gram model extends the skip-gram neural network model used in word2vec [21] and fastText tools [29]. fastText is a successful reimplementation of word2vec which is designed to compute the vector of each word using its neighbors. The extension we propose in the MeSH-gram model replaces the neighbors by the MeSH descriptors of the abstract where each word occurs

## Related Works

Several semantic similarity and relatedness measures have been proposed the last decades [27]. Many of them have been implemented in the UMLS::Similarity package [30] avalaible in the UMLS (Unified Medical Language System). They differ on the method used: path-based, content-based, UMLS-based, corpus-based, and more recently, methods based on word vectors and concepts vectors. Path-based measures [7] use the hierarchical structure of a taxonomy to measure similarity: concepts close to each other are more similar. For instance, Sajadi et al. [31,32] developed a ranking algorithm based on Wikipedia graph metrics and used it to compare biomedical concepts. Content-based information measures [33,34] quantify the amount of information a concept provides: the more specific concepts have a greater amount of information content. Other approaches [35,36] use the entire UMLS (Unified Medical Language System) Metathesaurus® [37] in order to compare the context in the definition of the concept to quantify its relatedness.

Several methods are vector-based: the concepts are represented by vectors and the relatedness is usually estimated using the cosine similarity between them. In [36], the authors proposed to compute gloss vectors based second order co–occurrences trained on WordNet. In [38], the authors computed the cosine of two Latent Semantic Indexing concept vectors based on Pointwise Mutual Information association measure matrix. Recent vector-based methods use neural networks in order to compute concept vectors. The word2vec [21] tool was trained on different corpora: OSHUMED (by Sajadi et al. [32]), PubMed/MEDLINE (by Chui et al. [23]), PubMed Central (by Muneeb et al. [22], Chiu et al. [23], and Pakhomov et al. [24]), and CLINICAL-ALL by [24]. Following the approach used by De Vine et al. [39] on OSHUMED, Yu et al. [25] trained word2vec on Pubmed/MEDLINE transformed to UMLS concepts using the MetaMap indexing tool [40].

Other recent methods rely on word vectors. In their previous work, Yu et al. [41] retrofitted word vectors obtained by word2vec with hierarchichal information from the MeSH thesaurus. Recently, Henry et al. [26] compared different ways to combine word vectors in order to compute multi-word term vectors. The compared multi-word term aggregation method consists in the summation (avergaing) of component word vectors, creating concept vectors using the MetaMap indexing tool [40], and creating multi-word term vectors using the compoundify tool based on the UMLS Specialist Lexicon as glossary [xx]. More recently, Henry et al. [27] use association measures for estimating semantic similarity and relatedness between biomedical concepts on PubMed/MEDLINE transformed to UMLS concepts.

The best performance results were obtained by [25-27]. Their respective approach relies either on MetaMap in order to transform the text corpus into UMLS concepts or on additional external ressources such as the Specialist Lexicon.

The MeSH-gram model we propose in this study relies on a straightforward approach: it computes the word vectors by only using the MeSH descriptors that are already included in the PubMed/MEDLINE corpus. The extension we propose in the MeSH-gram model replaces the neighbors by the MeSH descriptors of the abstract where each word occurs.

In order to evaluate MeSH-gram, we use publicly available manually annotated corpora: two subsets from Mayo Clinic (MiniMayoSRS) of the MayoSRS (Mayo Semantic Relatedness Set) developed by Pakhomov et al. [42], and two from UMNSRS (The University of Minnesota Semantic Relatedness Set) developed by Pakhomov et al. [43]. MeSH-gram results first compared to skip-gram and are then compared to twenty existing solutions reported in [27], including the best ones [25-27]. The MeSH-gram model has several advantages: (i) it avoids considering uninformative and too frequent words; (ii) there are less MeSH descriptors than possible context words; and (iii) MeSH descriptors are manually assigned and curated, which assures the best quality of indexing.

# Methods

Neural network language models learn word vectors by either maximizing the probability of a word given the context, referred to as the CBOW (Continuous Bag Of Words) approach, or by maximizing the probability of the context given a word, referred to as the skip-gram approach.

**Skip-gram Word Embedding Model**

Given $w_1 w_2 ... w_n$ a text line of words $w_i$, the skip-gram model maximizes the following average log probability:

$$\frac{1}{2r} \sum_{i=1}^{2r} \sum_{-r \leq j \leq r, j \neq 0} \log p(w_{t+j} | w_t)$$

where $w_t$ is the target word, $w_{t+j}$ is the context, and $r$ is the context window radius. The context words surrounding the target term are determined by the context window radius $r$.

The probability of a context word $w_c$ given a target word $w_t$, is computed by:

$$p(w_c | w_t) = \frac{\exp(V_{w_c}^T V_{w_t})}{\sum_{w=1}^{N} \exp(V_w^T V_{w_t})}$$

where $N$ is the vocabulary size, and $V_w$ represents the vector of the word $w$.

**MeSH-gram word embedding model**

The MeSH-gram word-embedding model proposed in this paper extends the skip-gram neural network model used in word2vec [21] and fastText [29] tools: it uses MeSH descriptors that are already included in the PubMed/MEDLINE corpus to compute the word vectors.

Given $w_1 w_2 ... w_n$ the words of a PubMed/MEDLINE abstract, and $m_1 m_2 ... m_k$ the MeSH descriptors associated to this abstract, the MeSH-gram model maximizes the following average log probability:

$$\frac{1}{k} \sum_{i=1}^{k} \log p(m_i | w_t)$$

where $w_t$ is the target word and $m_i$ is a MeSH descriptor.

The probability of a context MeSH descriptor $m_c$ given a target word $w_t$, is computed by:

$$p(m_c|w_t) = \frac{exp(V_{m_c}^T V_{w_t})}{\sum_{m=1}^{M} exp(V_m^T V_{w_t})}$$

where $M$ is number of MeSH descriptors, and $V_m$ represents the vector of the MeSH descriptor $m$.

We have adapted fastText [29] in order to feed the neural network with pairs of word/MeSH descriptor.

**Vector Representation and Similarity Computation**

Using our MeSH-gram model and skip-gram model for comparison, we built word vectors of dimension 300. For the skip-gram model, we computed the vectors considering several window sizes $W$ of 2, 5, 10 and 25.

In order to quantify the relatedness of a pair of words, the cosine distance between the distributional context vectors of each word is used. In the case of a multi-word term, the vector is generated by computing the average of the component word vectors that compose the term. As an example, for the term "epilepsy attack", the vector $V_{epilepsy\_attack}$ will be computed as $V_{epilepsy\_attack}=(V_{epilepsy}+V_{attack})/2$ where $V_{epilepsy}$ and $V_{attack}$ represent the vector of each word 3epilepsy and Attack respectively. Rather than combining word vectors after construction, multi-word term vectors may be constructed directly from a preprocessed training corpus in which multi-word terms have been identified [27] otherwise this will involve huge cost in preprocessing and storage requirements.

**Training Corpus**

We used PubMed/MEDLINE corpus that contains the abstracts of each article and the associated MeSH descriptors (URL: ftp://ftp.ncbi.nlm.nih.gov/pubmed/). The corpus was parsed with *pubmed_parser*, a python XML parser (URL: https://github.com/titipata/pubmed_parser) for PubMed dataset. Each abstract was tokenized using *polyglot* (URL: https://github.com/aboSamoor/polyglot).

For the skip-gram word embedding model using fastText, we prepared a file composed by concatenated abstracts, in which each abstract is splitted into one sentence per line. For the MeSH-gram word embedding model, we generated a file in which each line consists of an abstract with its MeSH descriptors. We have adapted fastText in order to read each line of this file and feed the neural network with pairs of (word, MeSH descriptor).

**Gold Standard**

In order to compare the MeSH-gram word embedding model proposed in this study with other methods, we used two evaluation benchmarks: MiniMayoSRS [42] and UMNSRS [43]. MiniMayoSRS consists of 29 clinical term pairs. Two thirty pairs (66.67%) contain a multi-word term. The relatedness of each word pair is rated by medical coders and also by physicians. UMNSRS consists of 566 and 586 pairs medical pairs, for measuring similarity and relatedness respectively. The degree of association between terms in each dataset was rated by four medical residents from the University of Minnesota medical school. As suggested by Pakhomov et al. [43], we use a subset of the ratings consisting of 401 pairs for the similarity set and 430 pairs for the relatedness set. Twenty (4.99%) and seventeen (3.95%) of the term pairs contain multi-word terms for the similarity and relatedness subsets respectively. All these clinical terms correspond to UMLS concepts included in the Metathesaurus®.

The correlations between the generated relatedness scores and the human-assigned scores are calculated using Spearman's rank.

# Results

**Skip-gram Model versus MeSH-gram Model**

The results of the experiments are in Table 1 in which a comparison is performed between the results obtained with skip-gram model and those obtained by the MeSH-gram model using our modified version of fastText according to the four gold standards: MiniMayoSRS rated by physicians (MiniMayoSRS phys.), MiniMayoSRS rated by medical coders (MiniMayoSRS cod.), UMNSRS for similarity (UMNSRS Sim.) and UMNSRS for relatedness (UMNSRS Rel.).

*Table 1 – Spearman's rank correlations between human scores and computed similarities.*
*W: window size; n: nb of pairs.*

|  | MiniMayo | | UMNSRS | |
| --- | --- | --- | --- | --- |
|  | **Phys.** **n=29** | **Cod.** **n=29** | **Sim.** **n=380** | **Rel.** **n=397** |
| Skip-gram | | | | |
| W=02 | 0.740 | 0.757 | 0.679 | 0.529 |
| W=05 | 0.763 | 0.779 | 0.704 | 0.576 |
| W=10 | 0.776 | 0.789 | 0.716 | 0.589 |
| W=25 | 0.766 | 0.781 | 0.718 | 0.608 |
|  | | | | |
| MeSH-gram | **0.811** | **0.855** | **0.724**[*] | **0.643**[**] |
|  | | | [*]n=387 | [**]n=407 |

**MeSH-gram Model compared to Previous Works**

Table 2 gathers the results obtained by the MeSH-gram model we developed and twenty previous works' results. It allows a comparison between all the models and on the same gold standards (MiniMayoSRS and UMNSRS). Table 2 complete the table 12 given by Henry et al. [27].

# Discussion

As one can see in Table 1 for the skip-gram model, the more the window is extended, the more the results are improved on the UMNSRS gold standard. The best results of skip-gram are obtained with a window size $W=10$ for the MiniMayoSRS set. This suggests that word vectors are a better solution when we consider an important number of context words in the abstract. The best results are obtained with the MeSH-gram model that considers MeSH descriptors as context for each term, suggesting that MeSH descriptors catch the semantics of all the abstracts associated with it. We can conclude that taking MeSH descriptors instead of context words gives better results than considering a large window size: (i) bigger window size does not lead necessary to better results, and (ii) MeSH descriptors are fewer than context words (50 context words for window size W=25) leading also a reduced computation time.

The deeper comparison results with tewenty methods displayed in Table 2 confirm that the MeSH-gram model give comparable results with best previous work methods on the four gold standard datasets. While the methods (1) and (2) rely on the translation of PudMed/MEDLINE text data into ULMS concepts, and methods (4) and (5) require additional steps or resources such as compoundify tool (4) and MetaMapped MEDLINE corpus (5), the MeSH-gram model uses only the raw text corpus as input. The best previous works' results are obtained by the method (2) and then the method (3). However the method (2) is not recommended by the authors themselves as it uses concept expansion which requires additional computation cost without significantly increasing the performances for any dataset [27]. MeSH-gram is comparable to method (3) with better results on three datasets. All those results allow us to conclude that UMLS information used by the methods (1) to (5) is already contained in the MeSH descriptors available in the PudMed/MEDLINE corpus and used by the MeSH-gram model. Using MeSH descriptors as context is a good solution for datasets founded on UMLS concepts. However, MeSH-gram should be evaluated on other types of similarities such as BioSimVerb and BioSimLex [44].

*Table 2 – Spearman's rank correlations between human scores and computed similarities using MeSH-gram and previous works'methods. n: nb of pairs (inspired by [27]).*

|  | MiniMayo | | UMNSRS | |
|---|---|---|---|---|
|  | **Phys.** | **Cod.** | **Sim.** | **Rel.** |
| MeSH-gram | 0.81 (n=29) | **0.86** (n=29) | **0.72** (n=387) | **0.64** (n=407) |
| (1) Henry et al. [27]; recommended | **0.84** (n=29) | 0.81 (n=29) | 0.69 (n=392) | **0.64** (n=418) |
| *(2) Henry et al.[27]; not recommended* | *0.85 (n=29)* | *0.84 (n=29)* | *0.73 (n=392)* | *0.66 (n=418)* |
| (3) Henry et al. [26]; CBOW words | 0.82 (n=29) | 0.82 (n=29) | 0.69 (n=374) | 0.61 (n=396) |
| (4) Henry et al. [26]; CBOW compounds | 0.80 (n=29) | 0.78 (n=28) | 0.70 (n=373) | 0.65 (n=393) |
| (5) Henry et al. [26]; CBOW concepts | 0.77 (n=29) | 0.83 (n=29) | 0.73 (n=388) | 0.60 (n=413) |
| (6) Yu et al. [25]; narrow +other relations | -- | -- | 0.69 (n=526) | 0.62 (n=543) |
|  |  |  | 0.68 (n=418) | 0.63 (n=427) |
| (7) Yu et al. [25]; no lexicons | -- | -- | 0.64 (n=526) | 0.59 (n=543) |
|  |  |  | 0.63 (n=418) | 0.59 (n=427) |
| (8) Yu et al. [41] | 0.70 (n=25) | 0.67 (n=25) | -- | -- |
| (9) Sajadi et al. [32]; HITS similarity | 0.67 (n=29) | 0.72 (n=29) | 0.58 (n=566) | 0.51 (n=587) |
| (10) Sajadi et al. [32]; (word2vec OSHUMED+UMLS) | -- | -- | 0.39 (n=566) | 0.39 (n=587) |
| (11) Sajadi et al. [32] (word2vec on OSHUMED) | -- | -- | 0.26 (n=566) | 0.29 (n=587) |
| (12) Chui et al. [23] | -- | -- | 0.65 (n=n/a) | 0.60 (n=n/a) |
| (13) Pakhomov et al. [24] | -- | -- | 0.62 (n=449) | 0.58 (n=458) |
| (14) Muneeb et al. [25] | -- | -- | 0.52 (n=462) | 0.45 (n=465) |
| (15) Workman et al. [32] | 0.67 (n=29) | -- | -- | -- |
|  | 0.69 (n=25) |  |  |  |
| (16) Patawardhan and Pedersen [36] | 0.59 (n=29) | 0.58 (n=29) | 0.58 (n=387) | 0.45 (n=412) |
| (17) Lin [34] | 0.42 (n=26) | 0.53 (n=26) | 0.49 (n=340) | 0.29 (n=360) |
| (18) Resnik [33] | 0.34 (n=26) | 0.46 (n=26) | 0.49 (n=340) | 0.26 (n=360) |
| (19) Rada et al. [7] | 0.35 (n=26) | 0.44 (n=26) | 0.53 (n=340) | 0.29 (n=360) |
| (20) Lesk [35] | 0.52 (n=29) | 0.57 (n=29) | 0.50 (n=387) | 0.33 (n=412) |

## Conclusions

In this paper, we proposed a new method, MeSH-gram, to create distributional word vectors using MeSH descriptors as word context. We evaluated our results on four standard evaluation datasets, MiniMayoSRS Physicians, MiniMayoSRS Coders, UMNSRS tagged for relatedness, and UMNSRS tagged for similarity, and compared it against skip-gram model as a baseline and previous methods. All the obtained results of Spearman's rank correlations between human scores and computed similarities show that MeSH-gram (i) outperforms the skip-gram model, and (ii) is comparable to the best recent methods but that need more computation and aditional external resources.

In our future works, we plan to include in MeSH-gram the MeSH qualifiers affiliated to the descriptors in order to have a more precise semantic meaning (e.g. cancer/complications is more precise than cancer). A second step is to use fastText subwords and the evaluation of MeSH-gram for other kinds of silmilarities such as BioSimVerb and BioSimLex. MeSH-gram may also be used in other languages than English, for instance in French bibliographic corpora such as CISMeF [45], as well as in annaotated electronic health records.

**Corresponding author :**
Lina F. Soualmia, PhD, HdR
lina.soualmia@litislab.eu
Normandie Université, CURIB (LITIS),
25 Rue Lucien Tesnière,
76130 Mont-Saint-Aignan, FRANCE
+33 232 955 173